# Retrieval-Augmented Guardrails for AI-Drafted Patient-Portal Messages: Error Taxonomy Construction and Large-Scale Evaluation

Wenyuan Chen[1*], Fateme Nateghi Haredasht[1*], Kameron C Black[2], François Grolleau[1], Emily Alsentzer[3], Jonathan H Chen[1,2,4**], Stephen P Ma[2**]

[1]*Stanford Center for Biomedical Informatics Research, Stanford, CA, USA*

[2]*Department of Medicine, Stanford University, Stanford, CA, USA*

[3]*Department of Biomedical Data Science, Stanford University, Stanford, CA, USA*

[4]*Stanford Clinical Excellence Research Center, Stanford University, Stanford, CA, USA*

*Equal Contribution

**Equal Contribution

Email: wenyuanc@stanford.edu , fnateghi@stanford.edu , spma@stanford.edu

Asynchronous patient–clinician messaging via EHR portals is a growing source of clinician workload, prompting interest in large language models (LLMs) to assist with draft responses. However, LLM outputs may contain clinical inaccuracies, omissions, or tone mismatches, making robust evaluation essential. Our contributions are threefold: (1) we introduce a clinically grounded error ontology comprising 5 domains and 59 granular error codes, developed through inductive coding and expert adjudication; (2) we develop a retrieval-augmented evaluation pipeline (RAEC) that leverages semantically similar historical message–response pairs to improve judgment quality; and (3) we provide a two-stage prompting architecture using DSPy to enable scalable, interpretable, and hierarchical error detection. Our approach assesses the quality of drafts both in isolation and with reference to similar past message–response pairs retrieved from institutional archives. Using a two-stage DSPy pipeline, we compared baseline and reference-enhanced evaluations on over 1,500 patient messages. Retrieval context improved error identification in domains such as clinical completeness and workflow appropriateness. Human validation on 100 messages demonstrated superior agreement (concordance = 50% vs. 33%) and performance (F1 = 0.500 vs. 0.256) of context-enhanced labels vs. baseline, supporting the use of our RAEC pipeline as AI guardrails for patient messaging.

## 1. Introduction

Asynchronous, text-based communication between patients and clinicians via government-mandated secure portals has become an integral component of contemporary care delivery since its launch in the early 2000s,[1-4] improving access, continuity, and patient engagement across diverse health-care settings,[5-7] but staffing has not kept pace; delayed replies, clinician burnout, and safety risks are rising.[8] To cope with the accelerating volume of inbox messages—especially in resource-constrained clinics—health systems are experimenting with large language models (LLMs) that draft replies for clinician review .

Despite their promise, state-of-the-art LLMs produce clinically consequential errors in roughly one-third to one-half of outputs.[9-10] Current monitoring approaches predominantly rely on clinician oversight at the point of care and retrospective manual reviews by dedicated reviewers. Unfortunately, these strategies are neither scalable nor capable of preventing real-time patient harm.[11-12] Moreover, errors in generated clinical text remain a major obstacle to the broader adoption of generative AI technologies in healthcare. There is an urgent need for automated, real-time safety mechanisms to detect and mitigate these errors as they occur, ensuring safe and effective deployment in clinical practice.[13]

In response to these challenges, we developed a real-time, multi-agent framework called Retrieval-Augmented Error Checking (RAEC) to evaluate and explain potential errors in LLM-generated patient messages before they reach clinicians or patients. RAEC combines three core innovations: (1) a comprehensive, clinician-vetted error ontology; (2) retrieval of local historical message context to personalize error detection; and (3) a team of agentic LLM evaluators that classify and justify errors at inference time. We hypothesize that incorporating historical context from individual clinicians will improve the system's accuracy and specificity in identifying clinically meaningful errors.

To test this hypothesis, we benchmark RAEC's performance against a rigorously curated, human-labeled reference standard. The pipeline ingests de-identified clinician–patient message pairs, filters and re-ranks candidates based on structured metadata, and passes retrieved examples to LLM agents for classification and explanation. By providing a scalable safety layer for LLM-powered messaging, RAEC supports safer, more context-aware AI deployment, with the potential to reduce clinician workload and improve communication quality across diverse care settings.

## 2. Related Work

The current evaluation of generative artificial intelligence (AI) in clinical contexts has significant limitations. Traditional surface-level metrics such as BLEU and ROUGE do not capture the nuanced clinical implications or real-world applicability of AI-generated outputs. Human review, while more accurate, is not scalable for large datasets or routine clinical deployment. Modular large language model (LLM) "judge" systems like G-Eval and RubricEval have been introduced to assess outputs across multiple quality dimensions using specialized prompts, but these typically evaluate text in isolation and lack alignment with real-world clinical scenarios or local practice standards.[14-15]

Recent advances in retrieval-augmented generation (RAG) have demonstrated that grounding LLM outputs in external knowledge, such as clinical guidelines, medical textbooks, or up-to-date literature, can substantially improve factuality, reduce hallucinations, and enhance alignment with clinical best practices.[16] For example, RAG has been shown to improve factual accuracy and clinical alignment in specialty applications such as gastrointestinal symptom triage chatbots and orthopedic exam question answering.[17-18] These approaches allow LLMs to ground their responses in up-to-date, guideline-based information, reducing hallucinations and improving alignment with clinical best practices.[19-21]

However, the application of retrieval for post-hoc error checking—especially with personalization to local clinical practice—remains underexplored. Combining modular LLM agents with retrieval-augmented error checking that leverages locally relevant examples addresses this gap, enhancing both the accuracy and real-world applicability of AI-assisted clinical messaging. This approach is supported by recent systematic reviews, which emphasize the need for robust, context-aware, and standardized evaluation frameworks to ensure safe and effective LLM integration in clinical workflows.[21-22]

Our work addresses this gap in two ways. First, we construct a comprehensive, inductively derived error taxonomy specifically for AI-generated patient portal messages, enabling fine-grained and structured evaluation. Second, we introduce a novel Retrieval-Augmented Error Checking (RAEC) framework—not to improve generation, but to improve evaluation. Instead of supplying reference documents to help draft a response, our system retrieves five similar, previously answered patient-clinician message pairs from a large institutional archive. These exemplars are used as context for an LLM-based judge to assess whether the draft response contains omissions, inappropriate recommendations, or other clinical flaws. This grounding in local precedent significantly reshapes error detection patterns, redirecting attention from superficial factual discrepancies to actionable guidance and completeness, and offers a scalable path toward safer LLM integration in clinical communication.

## 3. Data

### 3.1. *Raw Corpus and Preprocessing*

We extracted 246,588 patient–clinician message threads from the Epic-based Stanford Health Care SHC) enterprise EHR system, spanning a 3-month period from October 2024 to January 2025. These threads were pulled from the secure patient portal infrastructure and represent all asynchronous outpatient communications during that period. Each thread record includes structured metadata (timestamps, department and provider identifiers) along with the full text of the patient's new inquiry, the initial LLM-generated draft, and the final clinician-edited reply.

To ensure high-quality comparisons between AI drafts and human-edited responses, we focused on message triplets (patient query, LLM draft, clinician reply) that clearly represented a single, discrete communication turn. We excluded system messages, administrative events, and cases with missing or NULL content in any of the three key fields. Duplicates were collapsed based on identical text/timestamp pairs, and the earliest valid message was retained. This yielded a clean working corpus of 146,681 unique message triplets (220,739 total messages) across 11 clinical specialties, including Family Medicine, Primary Care, Internal Medicine, Oncology, Hematology, Geriatrics, Gastroenterology, Express Care, Radiation Oncology, Sports Medicine, and Coordinated Care.

For downstream evaluation, we retained only variables essential for automated error labeling and validation: Patient Message, LLM Prompt, LLM Draft, Clinician Reply, Date Sent, Recipient Name, Message Sender, Department, Specialty, and a pseudonymized thread identifier. All extraneous metadata was removed to reduce privacy exposure and enhance reproducibility.

### 3.2. *Sampling Strategy and Dataset Splits*

From the cleaned corpus, we first extracted a stratified random sample of 1,000 messages, maintaining the same specialty distribution observed in the full dataset, and dedicated this set exclusively to the inductive coding procedure that generated the error taxonomy (see § 4.1. ). After removing these cases, we again performed stratified random sampling—preserving the specialty proportions—to select 1570 additional messages for automated evaluation by the LLM guardrail. Finally, we performed independent physician review on a 100-message subset of this evaluation cohort (using 1:1 sampling for messages with at least one error identified by automated LLM guardrail vs messages with no errors); the manually reviewed labels from this set serve as the ground truth references in all validation analyses.

## 4. Methods

### 4.1. *Construction of a Clinical Error Taxonomy*

For our initial seed error taxonomy, physicians and researchers with experience studying AI-generated draft replies to patient messages brainstormed common discrepancies observed in patient communication and organized the resulting items into an initial hierarchy. To refine and expand it, we had an LLM perform inductive coding on a stratified sample of 1,000 message triplets (patient query, LLM draft, and clinician-edited reply), maintaining specialty proportions from the full corpus (see § 3.2. ). For this process, we used the OpenAI o3-mini reasoning model in a looped prompt–response design: for each message, the model labeled errors using the current taxonomy and proposed new error codes when existing ones didn't fit. The inductive codes were then reviewed in conjunction with the original codes by a board-certified physician. Simple, duplicative codes were manually resolved whereas overly broad codes/new thematic elements prompted the generation of additional codes and extensive reorganization of subdomains. This process involved additional manual inductive coding of 100 example messages for refinement before the taxonomy was finalized.

### 4.2. *Baseline: LLM Guardrail (No Retrieval)*

We implemented a two-stage LLM-based guardrail system using Declarative Self-improving Python (DSPY),[23] a modular orchestration framework for prompt-driven pipelines. In the baseline configuration, the model receives the patient message, the corresponding LLM-generated draft response, and structured EHR metadata available at the time of response drafting (e.g., patient name, department, last note, thread history, etc), with no additional retrieved context provided.

In the first stage, the LLM functions as an error identifier. It determines whether any error is present in the draft response and if so, produces a short summary, along with a free-text explanation of its reasoning. If an error is identified, the system proceeds to the second stage, where the model classifies the error using a structured taxonomy. Specifically, the LLM is provided with the full set of domain, subdomain, and error code definitions, and it returns a structured JSON output that includes the selected error code(s), associated confidence scores, and justification.

### 4.3. *Enhanced: Retrieval-Augmented Guardrail (RAEC)*

In the enhanced configuration of our evaluation pipeline, we activate a retrieval-augmented mode, referred to as RAEC (Retrieval-Augmented Error Checking). In this setting, the LLM receives not only the patient message, draft response, and structured metadata, but also a curated set of up to five similar message–response pairs drawn from the historical archive of clinician-authored replies (see§ 4.3.1. to 4.3.3. ). These retrieved examples serve as contextual reference points to inform the LLM's assessment of whether the current draft aligns with how similar cases were handled by real clinicians in the past. Importantly, the model is not prompted to quote, imitate, or directly compared to these examples; instead, they are offered purely to ground the model's clinical reasoning in precedent.

The remainder of the pipeline remains identical to the baseline setup. The LLM still performs a two-stage evaluation: first, detecting the presence of errors, then assigning structured error codes based on the established taxonomy. By keeping the task structure fixed and only varying the availability of reference examples, we are able to isolate the impact of contextual grounding on the model's ability to identify safety-relevant flaws in its own output.

#### 4.3.1. *Message-Response Embeddings*

To support the retrieval-augmented evaluation, we embedded all 220,739 messages from 146,681 patient–clinician threads using the all-mpnet-base-v2 model from SentenceTransformers.[24] These 768-dimensional vectors were stored alongside structured metadata such as recipient name, department, and specialty, for filtering purposes, enabling scalable and low-latency retrieval.

#### 4.3.2. *Message-Response Retrieval*

To ensure relevance, the retrieval process filters candidates by matching physician, department, and specialty metadata, and then ranks them based on cosine similarity between the query patient message and the archived patient messages, using sentence embeddings derived from the all-mpnet-base-v2 model in the SentenceTransformers library.[24-25] Once a top match is selected, its corresponding clinician response is retrieved as part of the reference pair. Only the top five message–response pairs are presented to the LLM, formatted as background context.

#### 4.3.3. *Message-Response Retrieval Evaluation*

To evaluate this retrieval mechanism, a board-certified physician reviewed a random sample of 56 queries and their associated retrieved sets. For each query, the reviewer assigned a binary usefulness score to each set of the five retrieved examples and provided an ordinal re-ranking of their relative clinical helpfulness. Retrieval quality was quantified via two metrics: Mean Usefulness (average fraction of retrieved messages marked helpful per query) and Kendall's τ (correlation between similarity-based and physician rankings).

### 4.4. *Physician Validation of Assigned Error Labels*

To rigorously assess the accuracy of the LLM-generated error labels, and assess the value of the retrieval augmentation process, we conducted a validation study using a subset of 100 messages from the evaluation cohort. Each patient message and corresponding LLM draft along with all available EHR context was reviewed by a board-certified physician. Using the finalized error taxonomy, they identified and labeled all applicable errors. These physician annotations were then treated as the reference standard against which both the baseline and retrieval-augmented guardrail outputs were evaluated.

For our primary outcome, we define message-level concordance as the absolute agreement between the LLM-based annotations (Baseline and Enhanced) with the Physician annotations at the various hierarchical levels of the taxonomy (domain, subdomain, or error code). More precisely, let $V(x)$ be the set of labels assigned to index $x$ by a source at a given level (e.g., domains assigned to index $x$). Then for each index and hierarchical level, concordance is defined as:

- *Concordant* = 1, if $V_{physician}(x) == V_{llm}(x)$
- *Concordant* = 0, otherwise

We computed counts and percentages of concordant cases for both Baseline and Enhanced guardrails and performed a statistical comparison of the two using McNemar's test.

We then computed standard performance metrics for all individual labels at all three hierarchical levels including sensitivity (recall), specificity, positive predictive value (precision), negative predictive value, and F1-score (harmonic mean of precision and recall). The denominator for true negatives included all index-label combinations including those not labeled by any source.

## 5. Results

### 5.1. *Inductive Coding and Error Taxonomy*

The tree diagram in Figure 1 traces the trajectory of the taxonomy across three phases. The physician seeded draft began with 6 domains, 27 subdomains, and 37 error codes. While the o3mini inductive loop annotated the stratified 1 000 message samples, it proposed 13 additional codes. After human review (see § 4.1), our final taxonomy resulted in 5 domains, 24 subdomains, and 59 error codes.

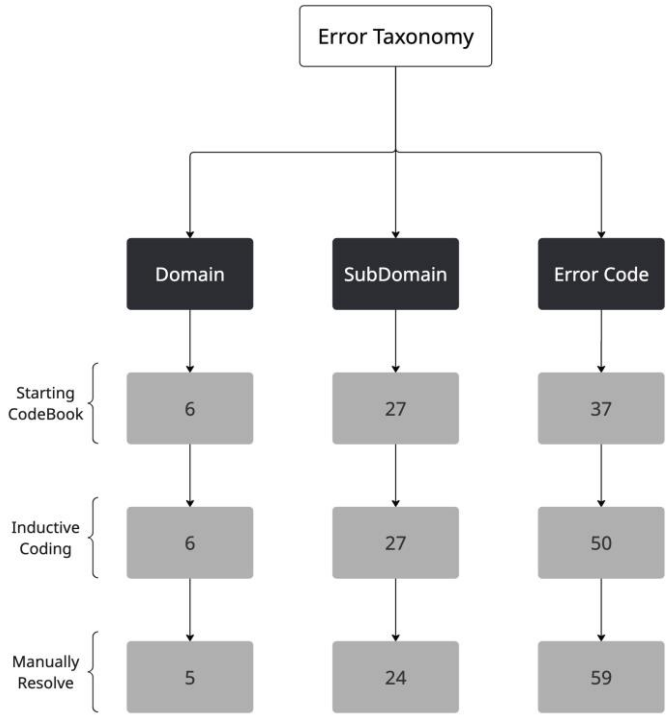

Figure 1: Schematic of error taxonomy development and refinement.

## 5.2. Reference Retrieval Performance

A board-certified physician adjudicated a random sample of 56 queries (280 retrieved message–response pairs). On average, 85.7 % of the queries returned clinically helpful references (Mean Usefulness = 0.86). The average Kendall's $\tau$ of 0.77 confirms strong agreement between the similarity-based ranking and the physician's preferred ordering, demonstrating that the retrieval stage both identifies and prioritizes exemplars likely to aid downstream error detection.

## 5.3. Physician Validation and Comparison of Baseline vs Enhanced Guardrails

### 5.3.1. Error Counts

In terms of basic summary counts (Table 1) for the 100 unique message indices evaluated using all three workflows (Baseline, Enhanced, and Physician), we found that Physicians flagged errors in more cases overall compared to the baseline and enhanced LLM judges (57 cases vs. 43 and 36, respectively). At a domain level, physicians tended to flag more errors in clinical reasoning and fewer errors in communication quality and readability. At the level of individual subdomains and error codes, the major changes occurred at the level of workflow recommendations (large increase in workflow violations) and communication clarity (large decrease in messages flagged as ambiguous or conflicting instructions). The other three error domains occurred infrequently, but physicians generally flagged fewer errors in those areas as well.

Table 1: Comparison of error counts identified by Physicians vs Baseline and Enhanced guardrails.

| Category | Baseline | Enhanced | Physician |
|---|---|---|---|
| **Cases with ≥1 error** | 43 | 36 | 57 |
| **Error rate (%)** | 43.0% | 36.0% | 57.0% |
| **Domain-level error counts** | | | |
| Accessibility | 1 | 0 | 1 |
| Bias & Stigmatization | 5 | 4 | 2 |
| Clinical Reasoning | 43 | 36 | 54 |
| Communication Quality & Readability | 37 | 32 | 16 |
| Privacy & Security | 4 | 7 | 0 |

### 5.3.2. *Message-Level Concordance*

The Enhanced guardrails were more frequently concordant with the Physician labels than the Baseline guardrails at the level of domains, subdomains, and error codes (Table 2).

Table 2: Comparison of Physician concordance rates between Baseline and Enhanced guardrails. Statistical significance was calculated using McNemar's test.

| Level | Baseline Concordance | Enhanced Concordance | P-value |
|---|---|---|---|
| **Domains** | 45 (45.0%) | 56 (56.0%) | 0.0371 |
| **Subdomains** | 35 (35.0%) | 50 (50.0%) | 0.0007 |
| **Error Codes** | 33 (33.0%) | 50 (50.0%) | 0.0002 |

We also perform subgroup analysis for individual domains, subdomains, and error codes (data not shown), with generally improved concordance for the Enhanced guardrails compared to the Baseline guardrails.

### 5.3.3. *Error-Level Performance*

Table 3: Performance metrics for Baseline and Enhanced guardrails at identification of the correct domains, subdomains, and error codes

| | Domain Level | | Subdomain Level | | Error Code Level | |
|---|---|---|---|---|---|---|
| | Baseline | Enhanced | Baseline | Enhanced | Baseline | Enhanced |
| **TP** | 42 | 47 | 36 | 55 | 30 | 59 |
| **FP** | 48 | 32 | 86 | 58 | 102 | 75 |
| **FN** | 31 | 26 | 58 | 39 | 72 | 43 |
| **TN** | 379 | 395 | 1420 | 1448 | 3496 | 3423 |
| **Sensitivity** | 0.575 | 0.644 | 0.383 | 0.585 | 0.294 | 0.578 |
| **Specificity** | 0.888 | 0.925 | 0.943 | 0.961 | 0.972 | 0.979 |
| **PPV** | 0.467 | 0.595 | 0.295 | 0.487 | 0.227 | 0.44 |
| **NPV** | 0.924 | 0.938 | 0.961 | 0.974 | 0.98 | 0.988 |
| **Accuracy** | 0.842 | 0.884 | 0.91 | 0.939 | 0.953 | 0.967 |
| **F1** | 0.515 | 0.618 | 0.333 | 0.531 | 0.256 | 0.500 |

Next, we measured the overall performance of our guardrails at identifying the correct domains, subdomains, and error codes (Table 3). In comparison to the Baseline guardrails where both precision (PPV) and recall (Sensitivity) drop significantly (0.467 to 0.227 and 0.575 to 0.294, respectively) as the granularity increases from domain to error code, the Enhanced guardrails maintain much better performance (0.595 to 0.44 and 0.644 to 0.578, respectively). This is also reflected in the gap between their F1 scores as granularity increases (0.618 vs 0.515 at the domain level, increasing to 0.500 vs. 0.256 at the error code level).

### 5.3.4. *Impact of Enhanced Guardrails compared to Baseline*

We then compared error identification between baseline and retrieval-enhanced guardrail configurations across the full 1570 message cases to better determine the potential impact on downstream clinical workflows between the two models if they were to be deployed into production.

The baseline model identified 350 cases with at least one associated error, whereas the enhanced retrieval-augmented guardrail reduced this count to 307, reflecting our prior observations about increased specificity. This was also reflected at the level of individual errors, where the baseline model flagged a total of 1,170 errors, while the retrieval-augmented guardrail identified 1,025.

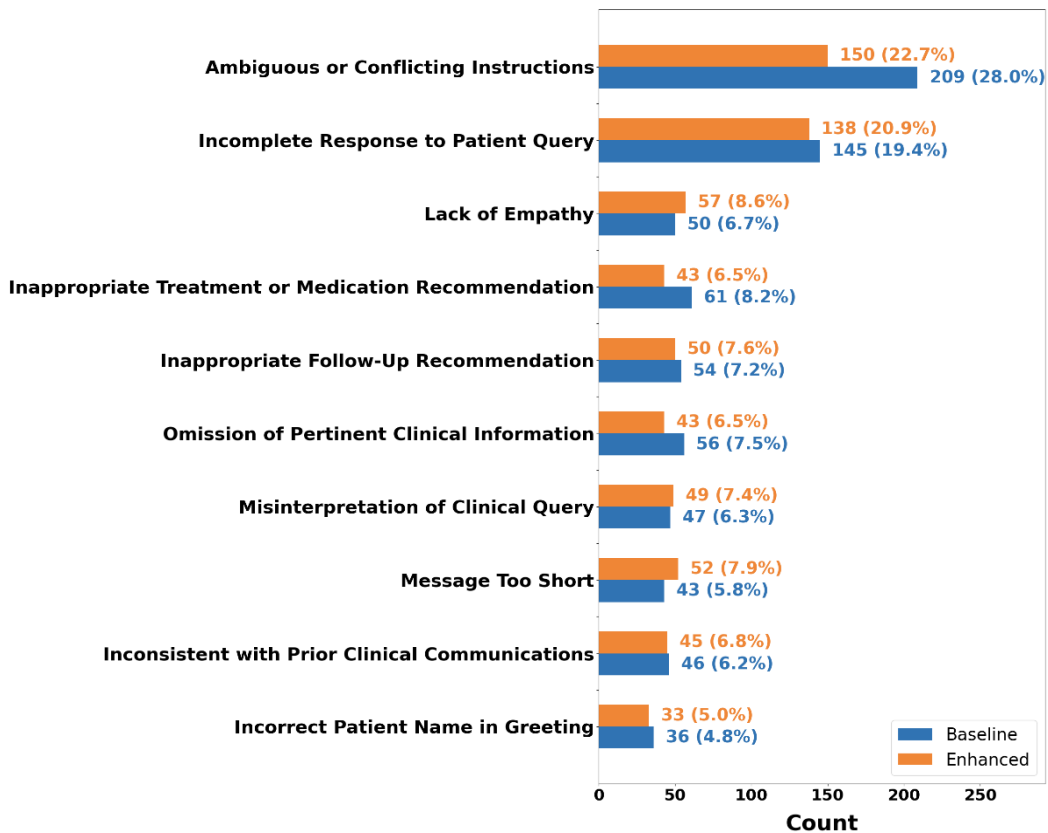

Figure 2: Relative frequencies of the top 10 most common error codes in baseline and enhanced models

When comparing error distributions (Figure 2), the Enhanced guardrail dramatically decreases the number of messages flagged as *Ambiguous or Conflicting Instructions* (28% to 22.7%). Out of the other 10 most common error codes, there were small increases in *Lack of Empathy* (6.7% to 8.6%), *Misinterpretation of Clinical Query* (6.3% to 7.4%), and *Message Too Short* (5.8% to 7.9%). The other error codes all showed small decreases, consistent with the overall increase in specificity for the Enhanced guardrail.

### 5.4. *Adapted vs Nonadapted Drafts*

Finally, to further validate our pipeline and demonstrate its utility, we utilize the Enhanced guardrails to analyze differences in the frequency of errors between messages where the drafts were used vs. messages where the drafts were not used for the same overall set of 1570 messages. We identified a decrease in the error rate for utilized drafts (36 errors across 132 messages = 0.27 errors per draft) vs. discarded drafts (989 errors across 1438 messages = 0.69 errors per draft).

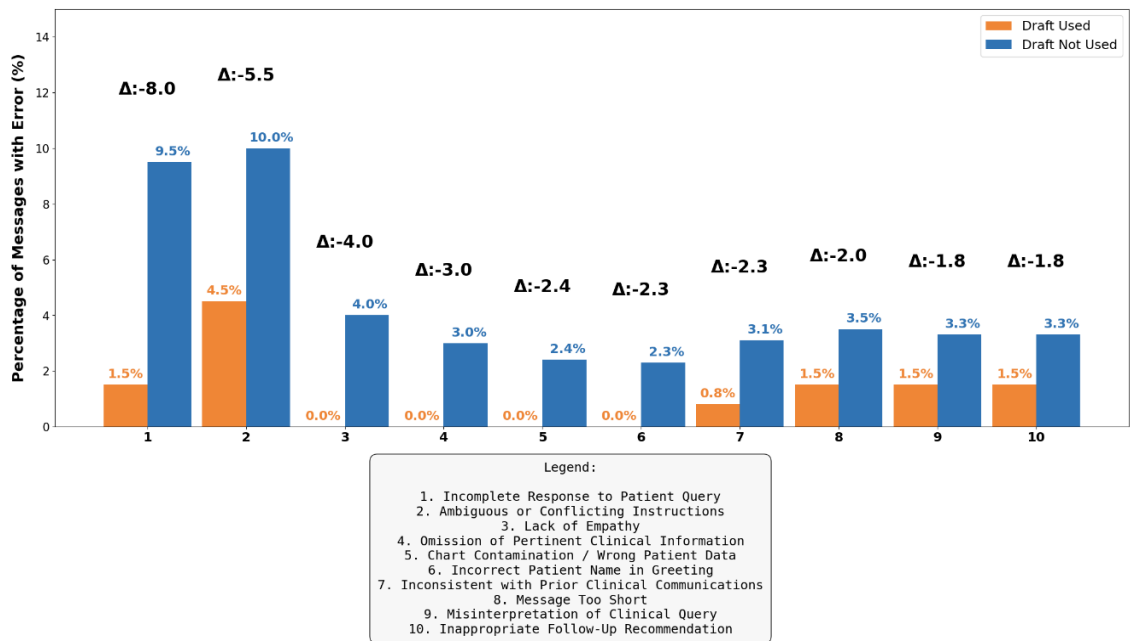

Figure 3: Error codes with the largest absolute drop in frequency for utilized drafts

We also ranked the individual error codes by their absolute change in frequency comparing utilized drafts vs. discarded drafts. The labels with the largest drop in frequency for utilized drafts (Figure 3) included instances where the reply to the patient was clearly incorrect (*Incomplete Response to Patient Query* at -8.0 pp, *Chart Contamination/Wrong Patient Data* at -2.4 pp, *Incorrect Patient Name in Greeting* at -2.3 pp) or lacked empathy (*Lack of Empathy* at *-4.0 pp*).

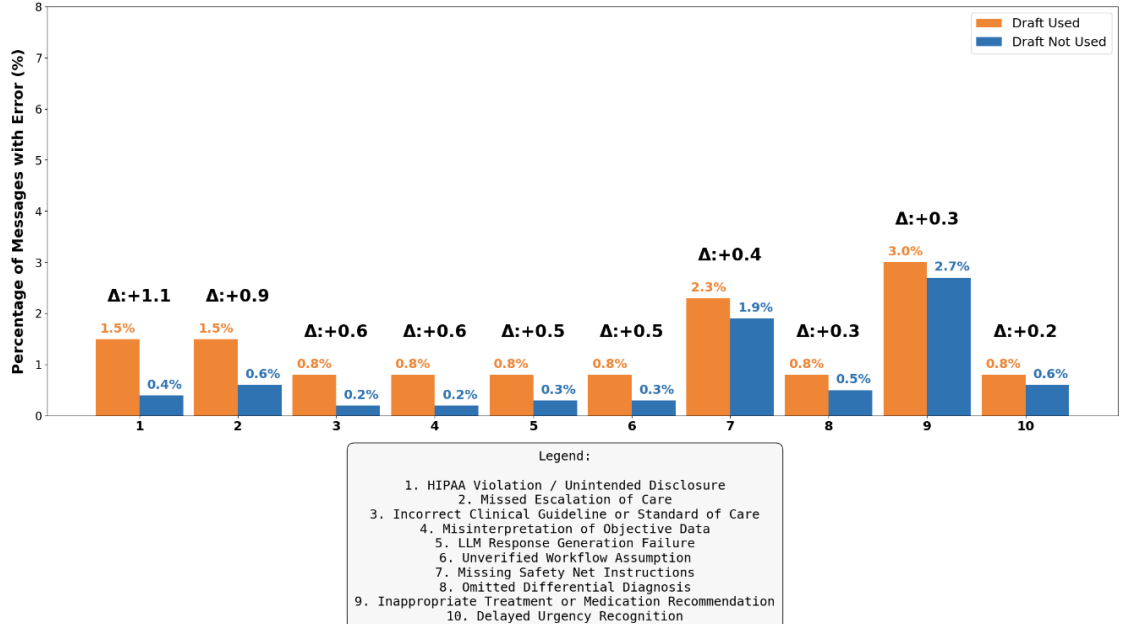

Figure 4: Error codes with the largest absolute increase in frequency for utilized drafts

In contrast, the error codes that were more common when drafts were used (Figure 4) included several subtle clinical errors including errors of omission (*Missed Escalation of Care* at +1.1 pp, *Incorrect Clinical Guideline or Standard of Care* at +0.9 pp, *Missed Safety Net Instructions* at +0.4 pp, *Omitted Differential Diagnosis* at +0.3).

## 6. Discussion

This study introduced and rigorously evaluated our Retrieval Augmented Error Checking (RAEC) pipeline, a modular framework for the evaluation of clinician replies to patient messages that synthesizes three separate innovations: 1) a comprehensive error taxonomy, 2) local context retrieval, and 3) an agentic two-stage LLM pipeline using DSPy to deliver real-time quality assurance for AI generated patient messages.

The lack of a comprehensive error taxonomy constitutes a major gap in the field for the evaluation of AI-generated draft replies, which currently relies on superficial metrics such as edit distances. We utilize machine-assisted inductive coding in conjunction with manual refinement to create an empirically grounded clinician validated taxonomy, allowing us to distinguish between superficial communication lapses vs. deeper clinical inaccuracies. Embedding this taxonomy within RAEC enabled nuanced, structured error detection with clear clinical interpretability.

We found that augmentation through local context retrieval was critically important for the performance of our LLM guardrails, with the Enhanced guardrail achieving much higher message-level concordance with Physician labels compared to the Baseline guardrail as well as much better performance for individual error detection across all traditional metrics. This improvement was most evident when identifying individual error codes but was present even when evaluating performance with less granularity at the level of general error domains. Importantly, we identified simultaneous improvements in both precision and recall meaning that in clinical practice, the Enhanced guardrail would simultaneously identify more errors (potentially preventing more harm to patients) while also decreasing the burden on clinicians (who will need to perform follow-up review of false positives). Subgroup analysis at the level of individual errors revealed that the use of Enhanced guardrails markedly reduced the total number of flagged messages and errors, with a major reduction in errors flagged as *Ambiguous or Conflicting Instructions*, which was noted to be an important source of false positives (see § 5.3.1. ). The use of retrieved context also resulted in slight increases in error counts for *Lack of Empathy*, and *Message Too Short*, which may reflect differences in clinician writing style in the retrieved context and has been identified as a major barrier to utilization of AI-generated draft replies.[26]

Despite this improvement in performance, we identified several opportunities for improvement based on the major shifts in error counts between the Enhanced guardrail labels and the Physician labels. Workflow violations were a notable blind spot for LLMs, which makes sense given that many of these workflows consist of tribal knowledge that are sometimes not even documented internally. While examples of these workflows were sometimes apparent in the retrieved reference pairs, they were not always detected by the Enhanced guardrails. On the other end of the spectrum, the LLMs frequently flagged messages as ambiguous or conflicting that were actually straightforward, which will likely require adjustments to the taxonomy definitions to adjust the threshold for meeting this error definition. In general, physicians more frequently flagged issues in clinical reasoning and less frequently identified problems related to communication quality and readability. This suggests that human reviewers may be more attuned to clinical nuances and interpretive gaps, while LLM-based systems may be better at identifying stylistic differences.

Finally, we further validated the performance of our pipeline and demonstrated additional utility by applying our Enhanced guardrails towards the analysis of clinician behavior through the comparison of error rates stratified by draft utilization. Our priori expectations were that better drafts would be more likely to be used by our clinicians and that obvious surface flaws would deter adoption whereas more subtle clinical reasoning errors might slip through. Our results confirmed our intuition, with a much lower overall error rate for utilized drafts (60% relative reduction from 0.69 errors/draft to 0.27 errors/draft), passing an essential sanity check for our pipeline. Moreover, we also identified relatively more errors related to subtle clinical reasoning when drafts were used, including missed escalations of care, safety net instructions, and considering a broader differential. These errors of omission are notoriously challenging to address as they require busy clinicians to go beyond directly replying to the patient message and spontaneously apply other frameworks/checklists.

## 7. Future Work

Despite these encouraging findings, the present work has important limitations that suggest clear avenues for refinement. First, the inductive coding phase drew on a corpus of only 1,000 message threads. Although this sample was adequate to capture the dominant error modes, it left several infrequent—but clinically meaningful—codes with sparse or no few shot exemplars. Enlarging the annotated corpus and engaging professional ontology engineers will be essential to expanding coverage of the "long tail" of rare errors. Second, our retrieval module relies on cosine similarity over allmpnetbasev2 sentence embeddings—a practical but generic choice. While this straightforward approach performed remarkably well at identifying and ranking relevant context as adjudicated by an independent physician, future iterations should explore domain adapted encoders, hybrid lexical–semantic rankers, or clinically informed rescoring schemes to surface even more contextually relevant exemplars. Third, we currently embed each patient message as a single text block for the purposes of retrieval. This strategy can obscure multiple clinical concerns that are often packed into a single thread. Tokenizing messages into topic coherent spans and retrieving context at sub-message granularity may further improve the guardrail's sensitivity to nuanced errors arising in complex, multi-issue communications.

## 8. Conclusion

Collectively, these findings establish RAEC as a robust, scalable solution for mitigating clinical risk in AI assisted messaging. By uniting a structured error taxonomy, local context retrieval, and modular LLM agents, the framework systematically produces contextually grounded judgments aligned with clinician expertise that elevates clinically salient errors while reducing noise—ultimately enhancing patient safety while alleviating the growing burden of asynchronous communication.